# Meat adulteration detection through digital image analysis of histological cuts using LBP

João J. de Macedo Neto, Jefersson A. dos Santos and William Robson Schwartz
*Department of Computer Science*
*Federal University of Minas Gerais, UFMG*
*Belo Horizonte, Brazil*
{joaomacedo, jefersson, william}@ufmg.br

**Abstract -** Food fraud has been an area of great concern due to its risk to public health, reduction of food quality or nutritional value and for its economic consequences. For this reason, it´s been object of regulation in many countries (e.g. [1], [2]). One type of food that has been frequently object of fraud through the addition of water or an aqueous solution is bovine meat. The traditional methods used to detect this kind of fraud are expensive, time-consuming and depend on physicochemical analysis that require complex laboratory techniques, specific for each added substance. In this paper, based on digital images of histological cuts of adulterated and not-adulterated (normal) bovine meat, we evaluate the of digital image analysis methods to identify the aforementioned kind of fraud, with focus on the Local Binary Pattern (LBP) algorithm.

## I. Introduction

Food fraud is a matter of major concern in many countries, not only for safety risks, but also for economic implications. According to the US Code, among many other reasons, food shall be deemed to be adulterated if any substance has been added thereto, increasing its bulk or weight, reducing its quality or strength, or makeing it appear better or of greater value than it is [2] without consent or knowledge of the customer.

One kind of food adulteration has been object of many police investigations by the Brazillian Federal Policy and recently has called public attention is bovine meat fraud. The adulteration process consists of injecting or adding water or an aqueous solution to the meat to increase its weight.

Traditional methods used for detecting of this kind of fraud require complex laboratory techniques that may vary according to the injected substance, making it expensive and time-consuming. To avoid these disadvantages, the development of new forensic methods to quickly detect the addition of exogenous substances in meat would be desirable.

Some researches based on image analysis of micrographic photo have already been suggested [3] and performed, but they depend of a human expert to analyze each individual image since the differences between these images are not always obvious. In fact, the differences can be very subtle in some cases.

The images used in this paper are from histological cuts of adulterated and unadulterated bovine meat, frozen at negative twenty Celsius degrees (-20ºC) with 20x amplification. From now on, we will refer to unadulterated meat as normal meat.

Figure 1 shows a sample image of histological cuts of normal bovine meat. Figures 2 and 3 show a sample image of histological cuts of adulterated bovine meat. As it was noticed by [3], the images of adulterated meat present disruption of the muscle tissue, with loss of normal contour and increased inter fiber space. In some cases (e.g. figure 2) it´s possible to see crystals with an amorphous aspect.

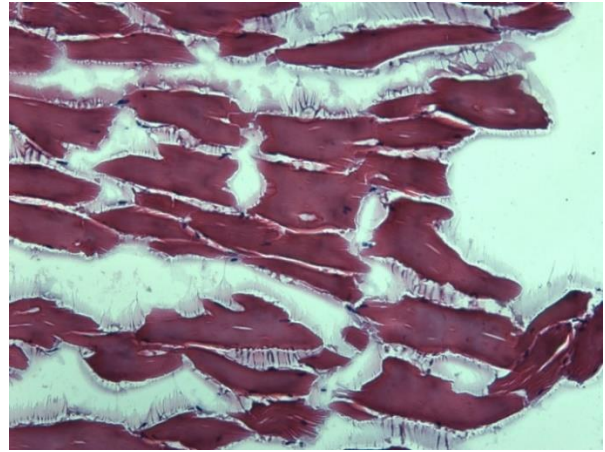

Fig. 1  Example of an image of histological cut of normal bovine meat

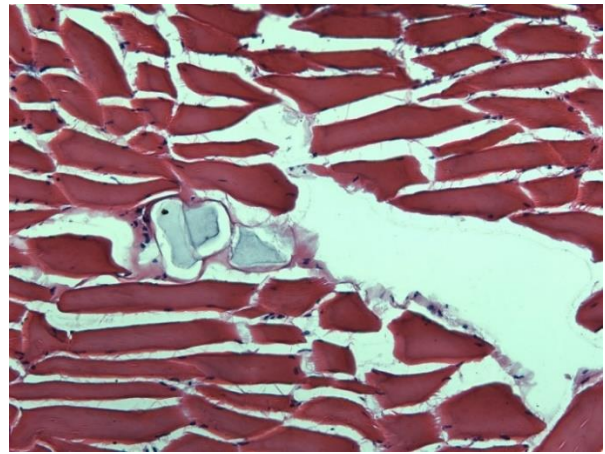

Fig. 2  Example of an image of histological cut of adulterated bovine meat (with noticeable visual indications of fraud)

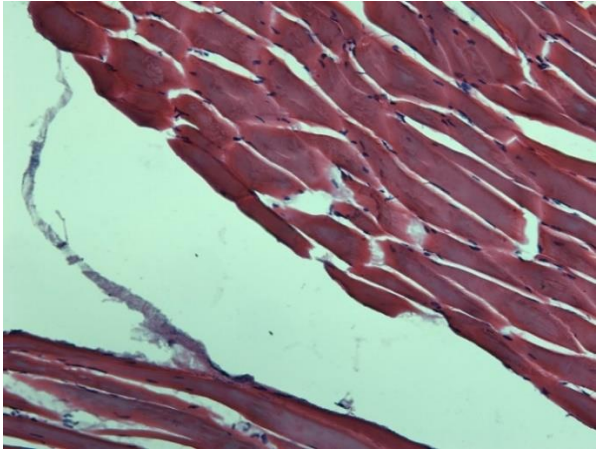

Fig. 3  Example of an image of histological cut of adulterated bovine meat (with subtle indications of fraud)

In this paper, we intend to evaluate the use of the Local Binary Pattern algorithm to analyze and classify digital images of histological cuts of normal and adulterated bovine meat as an automated technique to detect fraud in bovine meat through the injection or addition of water or an aqueous solution.

Based on a set of images of histological cuts of bovine meat divided in two classes (normal and adulterated) our objective is to extract the LBP histogram of these images and use them to train a classifier. This trained classifier will be used later to predict the class of new images of histological cuts, telling if they come from a normal or from an adulterated meat.

This work is divided into six sections. In section II, we make a brief description of the Local Binary Pattern method and we describe the feature extraction process. In section III, we describe the data set used to perform our experiments and the chosen methodology. In section IV, we describe our experiments. In section V we present and discuss the results of this work, and in the final section we make our conclusions and point some directions to follow in future works.

## II. LOCAL BINARY PATTERN

The Local Binary Pattern (LBP) algorithm was first introduced by Ojala et al. [4] as a method for image feature extraction to describe binary patterns in a texture. It´s a very simple, computational efficient and powerfull method that has been widely used for texture classification problems.

The method has many variants [5, 6, 7, 8], designed to implement new carachteristics not provided by the standard implementation. We used the cannonical LBP operator, that computes a value for each pixel in a gray-scale image by comparing its value with its eight nearest neighbors. Considering the pixel $g_c$ in figure 4:

| $g_0$ | $g_1$ | $g_2$ |
|---|---|---|
| $g_7$ | $g_c$ | $g_3$ |
| $g_6$ | $g_5$ | $g_4$ |

Fig. 4  $g_c$ (center pixel) and its neighbors to illustrate standard LBP operation

the LBP value will be computed by:

$$LBP(g_c) = \sum_{i=0}^{7} s(g_i - g_c).2^i \quad (1)$$

with

$$s(x) = \{1, x \geq 0;\ 0, x < 0 \quad (2)$$

and the resulting value will be a number between 0 and 255. The following image illustrates how (1) and (2) are used to compute a LBP value of one pixel of an image:

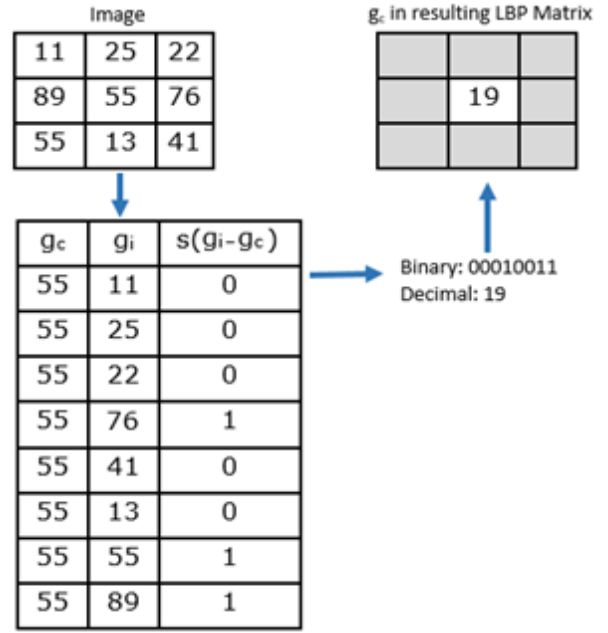

Fig. 5 LBP operator

This procedure is performed for each pixel in the original image, except for the borders, and a matrix is created with the resulting LBP values.

The code to compute (1) and (2) in an image is quite simple, and an example of its implementation written in C++ and using OpenCV is presented in figure 6.

```
lbp = cv::Mat::zeros(imagem.rows-2, imagem.cols-2, CV_8UC1);
for(int i=1;i<imagem.rows-1;i++) {
    for(int j=1;j<imagem.cols-1;j++) {
        unsigned char center = imagem.at<unsigned char>(i,j);
        unsigned char code = 0;
        code |= (imagem.at<unsigned char>(i-1,j-1) > center) << 7;
        code |= (imagem.at<unsigned char>(i-1,j)   > center) << 6;
        code |= (imagem.at<unsigned char>(i-1,j+1) > center) << 5;
        code |= (imagem.at<unsigned char>(i,j+1)   > center) << 4;
        code |= (imagem.at<unsigned char>(i+1,j+1) > center) << 3;
        code |= (imagem.at<unsigned char>(i+1,j)   > center) << 2;
        code |= (imagem.at<unsigned char>(i+1,j-1) > center) << 1;
        code |= (imagem.at<unsigned char>(i,j-1)   > center) << 0;
        lbp.at<unsigned char>(i-1,j-1) = code;
    }
}
```

Fig. 6  LBP implementation (C++/OpenCV)

The next step in using LBP consists in computing a histogram of the resulting LBP matrix. In the present work, this histogram (a vector of 256 numbers) will be used to train our classifier to learn the patterns of the images of histological cuts of normal and adulterated bovine meat.

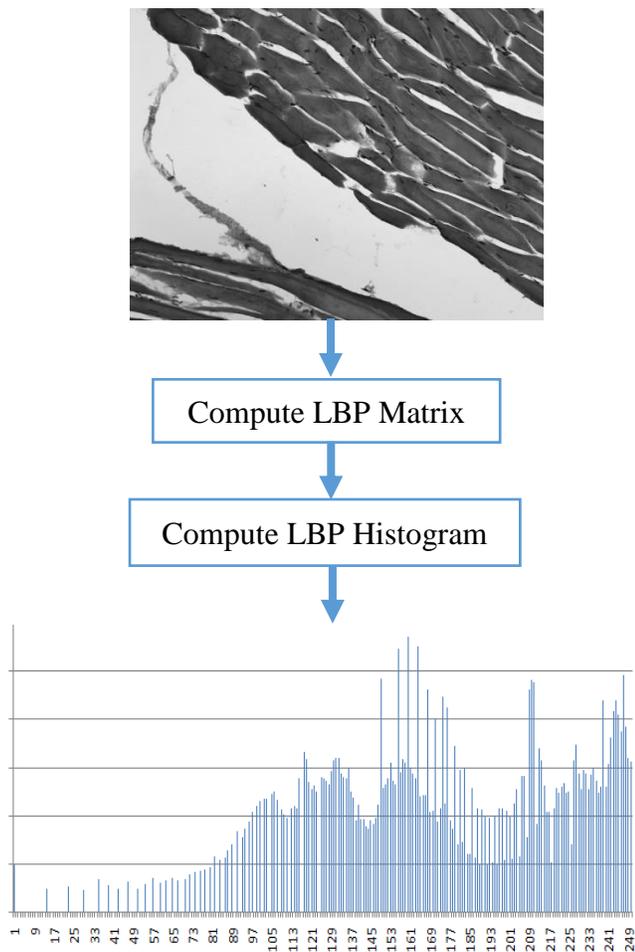

Fig. 7 LBP feature extraction cycle

The whole cycle of feature extraction using standard LBP is shown in figure 7. The image is shown in gray-scale because we are working with standard LBP, and our images need to be converted to gray-scale before the feature extraction process.

Before continuing in the next section, a few words about the LBP algorithm are necessary. Although the largescale use of the LBP descriptor in computer vision tasks has yielded good results, the method has known limitations and disadvantages, such as being sensible to noise and to image rotations, among others[9]. The sensibility to rotation can produce negative impacts in our experiments, since the micrographic photos of meat, especially the longitudinal histological cuts similar to that shown in figure 3, can produce negative impacts in the performance of our classifier. If this happens, a possible workaround for this problem is having a great amount of images, taken with different rotations. Another alternative is searching for a LBP variant thad already addresses this limitation of the standard method.

## III. DATA SET AND METHODOLOGY

This section describes the dataset used in this work, as well the methodology used to assess the generalization of our model and the classifier used for learning patterns and making predictions.

### A. Dataset

The dataset, provided by the Forensic Service of the Brazilian Federal Police, consists of images from frozen histological cuts of adulterated and unadulterated (normal) bovine meat. The histological cuts were produced from samples of frozen meat at negative twenty Celsius degrees (-20ºC), and the micrographic photos were taken with 20x amplification.

The dataset comes from two groups of images of histological cuts of normal and adulterated meat (table 1). The first group has 20 images of normal meat and 20 images of adulterated meat. The second group has 4 images of normal meat and 15 images of adulterated meat (this group is unbalanced). The two groups of images together makes a total of 59 images: 24 from normal meat and 35 from adulterated meat (unbalanced).

TABLE I
DATASET

|  | Normal | Adulterated | Total |
|---|---|---|---|
| **Group 1** | 20 | 20 | 40 |
| **Group 2** | 4 | 15 | 19 |
| **Total** | 24 | 35 | **59** |

### B. Methodology

Since there are only 59 images in the dataset, we decided to adopt the leave-one-out crossvalidation strategy: the classifier was trained with all images, except one that was left out for validation. This process was repeated 59 times, always leaving out one different image.

This same strategy was applied to the 40 images of the first group of the dataset in order to verify the results (with a smaller, but balanced version of our dataset).

### C. Classifier

Throughout our tests, it was used a linear Support Vector Machine (SVM) classifier for training and making predictions of our work.

We used the SVM provided by the machine learning package of the OpenCV framework, with the following parameters:

TABLE III
SVM CONFIGURATION/PARAMETERS

| Kernel type | Linear |
|---|---|
| **Type of SVM** | C-Support Vector Classification |
| **Termination criteria** | 100 iterations (max.) accuracy $10^{-6}$ |

## IV. EXPERIMENTS

Our meat fraud detection method, based on images of histological cuts of normal and adulterated bovine meat, was validated on the dataset described in section III and on a smaller balanced version containing the images from the first group of images (20 of each class), as described in table I.

In order to assess the importance of color information to the determination of the classes of images in the dataset and to have something to compare with the results of the LBP histogram, we also computed a color histogram of the gray-scale images and submitted them to our classifier (using the leave-one-out crossvalidation strategy). Besides that, we concatenated these two histograms (LBP + Color) and used it in our tests to evaluate whether this combination would produce better results.

In the first tests, we noticed that working with the images in their original resolution usually produced worse results than working with the images in smaller resolutions. We intuitively tought that it was due to the nature of the LBP algorithm, that compares one pixel with its neighbors, so when we deal with greater resolutions, the pixel variations are sparser and the extracted LBP descriptor becomes less discriminant. It was also noticed that the results were not good with very small resolutions.

To find a resolution with better results, we empirically evaluated the performance of the classifier using the whole dataset with multiple resolutions, stretching the images proportionally in a wide range of resolutions.

The best results using were verified in the resolutions from 50x37 to 300x225 (see figure 8). The same test was executed with the 40 images from group 1 of the dataset and we had similar results. Although the best performances in figure 8 were reached at resolutions 125x94 and 275x207, we chose the resolution 300x225 to continue the experiments.

## V. Results and Discussion

Figure 9 shows a more detailed view of our classifier´s acuracy using LBP, Color and LBP+Color histograms on the dataset (unbalanced).

The use of the LBP method to classify the images of the dataset yielded good results, what didn´t happen with the use of the color histogram. The third column in figure 9 shows the accuracy of the classifier using the concatenation of the LBP and color histograms, and we noticed that this strategy, in general, didn´t improve the performance of the classifier using only the LBP histogram.

Since we have an unbalanced dataset (with 24 images of normal meat and 35 images of adulterated meat) and in figure 9 we only show the global accuracy of the classifiers, we found it would be usefull to show the accuracy of the classifiers separated by classes (figure 10). We noticed that the per-class accuracy of the classifier using the LBP histogram was not much different from the global acuracy shown in figure 9 (the classifier trained with the LBP descriptor made one incorrect prediction for each).

We analyzed the images that the classifier (LBP) incorrectly predicted: the first one is an image of normal (unadulterated) meat that belongs to the second group of the dataset, which contains 4 images of normal meat and 15 images of adulterated meat; the second misclassified image is of adulterated meat and belongs to the first group of the dataset.

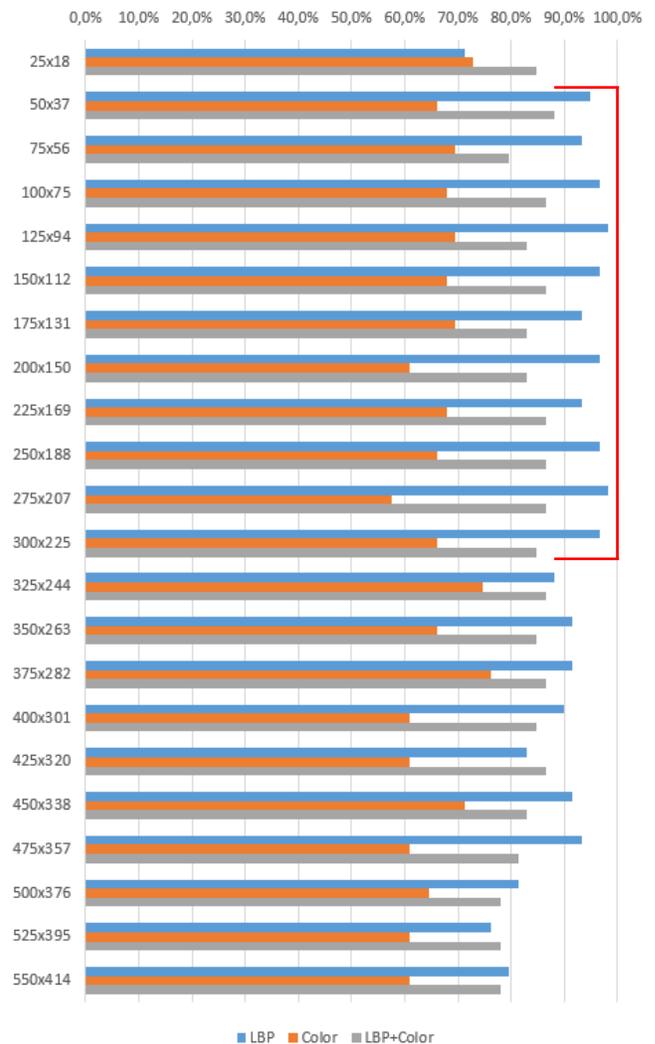

Fig. 8 Classifier´s acuracy using LBP, Color and LBB+Color histograms extracted from images of different resolutions

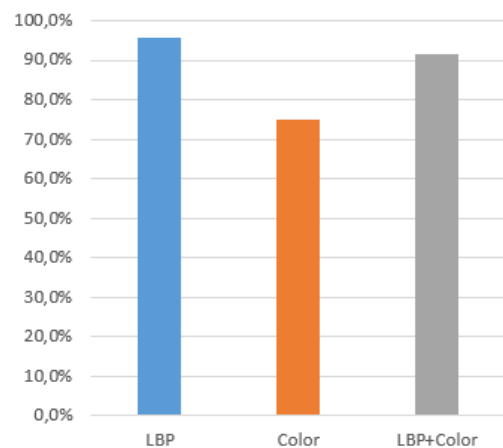

| LBP | Color | LBP+Color |
|---|---|---|
| 96,6% | 62,7% | 89,8% |

Fig. 9 Classifier´s acuracy using LBP, Color and LBB+Color histograms

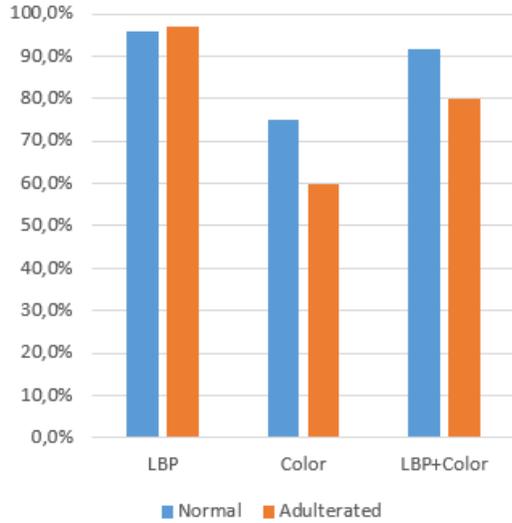

Fig. 10 Classifier´s acuracy using LBP, Color and LBB+Color histograms separated by classes

| Class | LBP | Color | LBP+Color |
|---|---|---|---|
| Normal | 95,8% | 75,0% | 91,7% |
| Adulterated | 97,1% | 60,0% | 80,0% |

The same tests were executed with a smaller version of the dataset, containing only the images of the first group (table 1), where there are 20 images of normal meat and 20 images of adulterated meat. Although this is a smaller dataset, it has the advantage of beign balanced. The classifier´s acuracy using LBP, Color and LBB+Color histograms on this smaller version of the dataset is shown in figure 11 and 12 (separated by classes in this one).

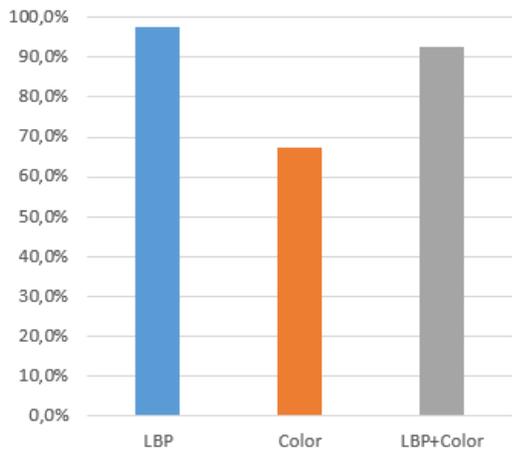

| LBP | Color | LBP+Color |
|---|---|---|
| 97,5% | 67,5% | 92,5% |

Fig. 11 Classifier´s acuracy using LBP, Color and LBB+Color histograms separated by classes on a smaller version of the dataset (with 40 images - balanced)

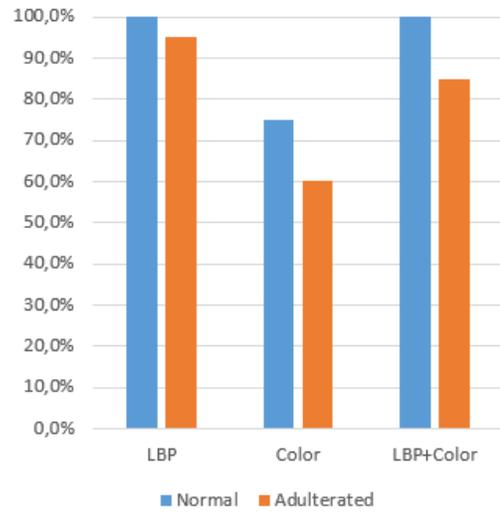

| Class | LBP | Color | LBP+Color |
|---|---|---|---|
| Normal | 100,0% | 75,0% | 100,0% |
| Adulterated | 95,0% | 60,0% | 85,0% |

Fig. 12 Classifier´s acuracy using LBP, Color and LBB+Color histograms separated by classes on a smaller version of the dataset (with 40 images - balanced)

The classifier trained with the LBP descriptor made only one mistake. The misclassified image in this smaller version of the dataset is of adulterated meat (figure 12) and it is different from the misclassified image of adulterated in the experiment on the whole dataset (figures 9 and 10).

## VI. CONCLUSION

The development of automated detection methods for food fraud analysis is a relevant advance in the use of micrographic photos for this kind of investigation, supporting the human expert analysis and providing an effective alternative to traditional laboratory analysis methods, that are more expensive and time-consuming, and may be innefective if the injected substance is not known or found.

In this work, it was shown that it is possible to classify images of histological cuts of frozen meat in order to determine whether it suffered an adulteration process through the addition or injection of water or an aqueous solution.

The results were considered satisfactory, even without a great number of image samples. In our dataset of 59 images (unbalanced), using leave-one-out crossvalidation, the classifier reached an accuracy of 96,6% (95,8% for normal meat and 97,1% for adulterated meat) trained with the standard LBP descriptor extracted from images resized to 300x225. Using a reduced and balanced version of the dataset, with images from group 1 (see table I), the algorithm reached an accuracy of 97,5% (100% for normal meat and 97,1% for adulterated meat).

### A. Future Work

Due to the known limitations and disadvantages of the standard LBP method, such as being sensible to noise and to image rotations, that can produce negative impacts in our experiments, it would be desirable to test the method with other variants of the LBP method that already address these limitations. Another alternative would be the evaluation of different image feature extraction algorithms, making a baseline for comparison purposes.

It would also be desirable to run the tests with a more robust dataset, collected with the same requisites (micrographic photos of histological cuts of normal and adulterated bovine meat, frozen at negative twenty Celsius degrees, taken with 20x amplification), but from different research centers. The adulterated samples of bovine meat could also be generated by the addition or injection of different aqueous solutions. With these directives, we believe that we would have a more realistic scenario to evaluate the widespread use of this technique.

## ACKNOWLEDGMENT

We thank Antonio Maurício Pires, from the Forensic Service of the Brazilian Federal Police, for many useful comments and for providing the images from the two groups (table I) of the dataset used in this work.